# DSP: A Differential Spatial Prediction Scheme for Comprehensive real industrial datasets

Jun-Jie Zhang[1], Cong Zhang[2]*, Neal N. Xiong[3]

[1,2]School of Mathematics and Computer science, Wuhan Polytechnic University, Wuhan 430023, China
Corresponding author: Cong Zhang (hb_wh_zc@163.com)
[3] Department of Mathematics and Computer Science Northeastern State University, Tahlequah, OK, USA

**Abstract:** Inverse Distance Weighted models (IDW) have been widely used for predicting and modeling multidimensional space in multimodal industrial processes. However, the more complex the structure of multidimensional space, the lower the performance of IDW models, and real industrial datasets tend to have more complex spatial structure. To solve this problem, a new framework for spatial prediction and modeling based on deep reinforcement learning network is proposed. In the proposed framework, the internal relationship between state and action is enhanced by reusing the state values in the Q network, and the convergence rate and stability of the deep reinforcement learning network are improved. The improved deep reinforcement learning network is then used to search for and learn the hyperparameters of each sample point in the inverse distance weighted model. These hyperparameters can reflect the spatial structure of the current industrial dataset to some extent. Then a spatial distribution of hyperparameters is constructed based on the learned hyperparameters. Each interpolation point obtains corresponding hyperparameters from the hyperparametric spatial distribution and brings them into the classical IDW models for prediction, thus achieving differential spatial prediction and modeling. The simulation results show that the proposed framework is suitable for real industrial datasets with complex spatial structure characteristics and is more accurate than current IDW models in spatial prediction.

## 1. Introduction

Spatial prediction is at the heart of spatial and spatio-temporal statistics. Common objectives are to predict a spatial process at unobserved locations and to study the spatial dependence in the region of interest. An ideal spatial prediction not only provides point prediction, but also distributional information such as quantiles or the density function to quantify uncertainties, risks and extreme values [1]. Spatial prediction is widely used in the field of Geology and environmental science [2]. Due to the rise of interdisciplinary in recent years, spatial prediction has been extended to other fields, such as biological science, computer vision, economics and public health [3-6].

The widely used methods in spatial prediction include inverse distance weighted interpolation and ordinary Kriging interpolation [7]. When using ordinary Kriging interpolation for spatial prediction, it is generally assumed that the covariance function is stationary, but the physical process is often non-Gaussian and non-stationary. The spatial covariance varies across space, e.g., in urban versus rural areas [8]. For inverse distance weighting algorithm, the predicted value of the interpolation point is the weighted average of each sample point. The advantage is that the calculation rate is fast and the size of the interpolated dataset is not strictly required. But the interpolation process of the inverse distance weighted interpolation algorithm is not associated with any actual physical process. However, spatial prediction for more general spatial processes remains an open problem.

Recently, due to the rise of machine learning, deep learning and deep neural network have been widely used in spatial prediction and classification, especially in computer vision and natural language processing [9]. Deep neural networks are effective not only in the spatial features of linearity and stationarity, but also in the prediction of complex features such as non-linearity and non-stationarity and computationally efficient in analyzing massive datasets using GPUs [10] [11]. However, there is an obstacle when using deep learning to predict spatial data. Classical deep neural networks cannot directly contain spatial dependence. The application of deep neural network in spatial prediction usually only includes spatial coordinates as features [12]. Using these features for spatial prediction may not be enough. Recently, convolutional neural networks (CNNs) [13] [14] have been stated to successfully capture the spatial and temporal dependencies in image processing through the relevant filters. However, the framework is designed for applications with large feature space, and has strict requirements on data scale, so it is not suitable for many spatial prediction problems with only in-situ observation and sparse observation. Compared with CNN, the purpose of spatial prediction is to consider the spatial correlation of response variables with limited observation features and sparse observation.

Therefore, based on the inverse distance weighted algorithm, we combine the characteristics of deep reinforcement learning network with the hyperparameters of inverse distance weighted algorithm, and propose an algorithm which can realize differential spatial interpolation. The proposed method is suitable for non-Gaussian data and does not make any assumptions about the interpolation space before interpolation. The proposed method has at least five contributions to spatial prediction and deep learning

1）It improves the convergence rate and stability of the deep reinforcement learning network by reusing the state values in the Q network.
2）It establishes a direct link between deep reinforcement learning network and inverse distance weighting in spatial prediction.
3）It enhances the relationship between interpolation points and the interpolation environment by mapping the spatial structure of the current interpolation point using the optimal hyperparameter in the inverse distance weighted method.

4） It uses the self-regulated learning characteristic of the deep reinforcement learning network to model spatial dependence.
5） It uses different hyperparameter for different interpolation points to achieve differential spatial prediction

For the remainder of this paper, the organization is as follows. In Section 2, we present the existing spatial prediction schemes (IDW) and classic deep reinforcement learning network: DQN, DDQN, DuDQN. In Section 3, we discuss the RSV-DuDQN-IDW algorithm and develop the guidelines for assigning appropriate hyperparameters adaptively to improve system performance, based on the theoretic analysis. Section 4 presents simulation model and discusses the simulation results. Finally, we give more related work related with IDW in Section 5, and conclude our work and discuss further work in Section 6.

**2. System model and Definitions**

*2.1 Inverse distance weighting algorithm (IDW)*

It is a multivariate spatial interpolation method based on the principle of similarity [15]. Each sample point has some influence on the interpolation point, called weight. The weight decreases with the increase of the distance between the sampling point and the interpolation point, and the closer the sampling point is to the interpolation point, the greater the weight is. Moreover, when the sampling point is within a certain range from the interpolation point, the weight can be ignored. Its biggest advantage is simple calculation and fast interpolation speed. The inverse distance weighting algorithm is based on the reciprocal of the distance between the interpolation point and the sampling point or the reciprocal of the distance n (n > 0) power, and then take the weighted average value of all adjacent points.    For the estimated value Y of point P, its general form is [16]:

$$Y(x) = \frac{\sum_{i=1}^{n} w_i(x) Y_i}{\sum_{i=1}^{n} w_i(x)} \qquad (1)$$

$$w_i(x) = \frac{1}{d(x, x_i)^p}, \qquad (i = 0, 1, \ldots, \ N; p \in N) \quad (2)$$

x is the interpolation point, xi is the sampling point, Yi is the value of the given point xi, N is the total number of sampling points used for interpolation, D (x, xi) is the distance from sampling point xi to interpolation point x. The weight wi decreases with the increase of the distance from the interpolation point. The greater the P value, the closer the distance to the interpolation point, and the greater the influence on the value of the interpolation point.

*2.2 Reinforcement learning*

Reinforcement Learning (RL) is a machine Learning method formed by the combination of animal psychology, control theory and other related disciplines [17]. In the process of learning, the agent of reinforcement learning learns through trial and error and seeks for the strategy to obtain the largest cumulative reward in the current environment, that is, the best strategy of Markov's decision-making process [18]. At present, reinforcement learning has received close attention from the industry and researchers, and abundant research achievements have been made in optimization, control, simulation and other fields [19] [20].

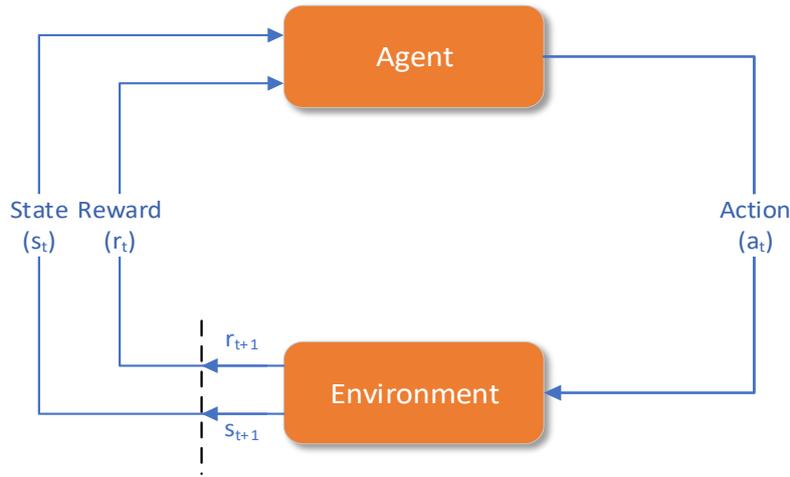

Figure 1. Framework of the Reinforcement Learning

The reinforcement learning framework is shown in Fig. 1. Under the current state st, the agent takes action at, and according to the state transfer function P, the environmental state will be transferred from st to st+1. At the same time, the environment will feed back a reward signal r to the agent according to the behavior at under the state st. The agent circulates this process for many times, aiming at obtaining the maximum accumulated reward. Through constant training, the optimal strategy of this process is finally obtained.

*2.3 Deep Q-Learning Network*

The deep Q-Learning network algorithm is a classical deep reinforcement learning algorithm, in which the deep learning part can perceive the environment information, while the enhanced learning part can make decisions based on the environment information provided by the deep learning part, complete the mapping from state to action, and get rewards. Then, the information is transformed into training data and provided to the deep learning to continuously optimize the neural network [21]. Deep Q-Learning network uses neural network to approximate Q-table value, but it also destroys the unconditional convergence of Q-Learning. To solve this problem, DQN has been improved from the following two aspects.

In the first aspect, the previous state is highly correlated with the current state in the iterative interaction between the DQN agent and the environment. If it is directly input into the neural network without processing, the neural network will overfit and fail to converge. Therefore, a memory buffer is added to DQN to store training samples over a period of time. During each learning process, DQN will randomly select a batch of samples from the memory buffer, input them into the deep neural network, and learn the gradient descent. When new training samples are generated, the old training samples and the new training samples are updated in mixed batches, thus interrupting the correlation between adjacent training samples and improving the utilization rate of training samples.

In the second aspect, a Q-Target neural network with the same structure as the current Q-Evaluate network but different parameters are established in DQN. This network is only used to calculate the Target Q value, while the current Q value is only predicted by the current Q-Evaluate network. This approach reduces the correlation between the target value and the current value. The loss function formula is computed by formula (3).

$$I = (r + \gamma \max Q_{a'}(s', a'; \omega^-)) - Q(s, a; \omega))^2 \qquad (3)$$

Where s represents the current state, a represents the action performed, and r represents the reward value of the environment to the agent. Q (s, a; $\omega$) is the output value of the current Q-Evaluate network to evaluate the value function of the current state-action pair when an action a is executed in the state of s. Q (s', a'; $\omega^-$) is the Q value of the target value function calculated using the Q-Target network.

The parameter $\omega$ of the Q-Evaluate network is updated in real time after each round of training, while the parameter $\omega^-$ of the Q-Target network is updated delayed, that is, all the parameters of the Q-Evaluate network are fully assigned to the Q-Target network after several rounds of training. The update formula of the value function can be obtained by solving the parameter $\omega$:

$$\omega_{t+1} = \omega_t + \alpha[r + \gamma \max Q_{a'}(s', a'; \omega^-) - Q(s, a; \omega)]\nabla Q(s, a; \omega) \quad (4)$$

## 2.4 Double Deep Q-Learning Network

When the classical reinforcement learning algorithms Q-Learning and deep Q-Learning network are used to make decisions and evaluate actions, the value of Q-max will be referred. Because the action selected according to the value of Q-max is not necessarily the action selected by the next state, it will lead to overestimation of Q-Target. To solve this problem, Van Hasselt proposed double deep Q learning, where DDQN eliminates the problem of Q-Target overestimation by decoupling the selection of Q-Estimate actions and the calculation of Q-Estimate [22].

DDQN and classical DQN also have two neural networks with the same structure [23], but their internal parameter updating has time difference, and their updating method is shown in the following formula (5).

$$Y_t = R_{t+1} + \gamma \max Q(S_{t+1}, \alpha; \theta_t) \qquad (5)$$

Since the Q-max predicted by the neural network has errors inherently and improves the neural network toward the Q-Target of the maximum error each time, the Q-max results in over estimation. Therefore, double-DQN introduces another neural network to optimize the error effect. There are two neural networks in the original DQN, Q-Evaluate and Q-Target. Q-evaluate neural network is used to predict the maximum action value of Q-max (S, A) in the Q-Target neural network, and then the Q value in q-Target is selected by this action. The update method is shown in the following formula (6).

$$Y_t = R_{t+1} + \gamma Q(S_{t+1}, \arg\max Q(S_{t+1}, \alpha; \theta_t), \theta_t) \qquad (6)$$

## 2.5 Dueling Deep Q-Learning Network

In DDQN, the algorithm is optimized by reducing the over-estimation of the target Q value, while in Dueling Deep Q-Learning Network (DuDQN) [24], the algorithm is optimized by improving the structure of the neural Network. DuDQN divides the Q network in a classic DQN into two parts. The first part is the value function part, which is only related to the state s and has nothing to do with the action to be taken, denoted as V(s; w, α); The second part is the advantage function part, which is not only related to the current state s, but also related to the action a to be performed, denoted as A(s, a; w, β). Therefore, the output of Q network in DuDQN is computed by formula (7).

$$Q(s,a;\theta,\alpha,\beta) = V(s;\theta,\beta) + A(s,a;\theta,\alpha) \quad (7)$$

Since the Q network of DuDQN outputs Q values directly, it is impossible to distinguish the roles of the value function part and the advantage function part. In order to reflect this identifiability, the formula is modified appropriately. Modified formula is shown in the following formula (8).

$$Q(s,a;\theta,\alpha,\beta) = V(s;\theta,\beta) + A(s,a;\theta,\alpha) - \max_{a' \in A} A(s,a';\theta,\alpha) \quad (8)$$

In practical application, the mean value of dominant function is usually used to replace the maximum value of advantage function, which improves the stability of optimization to some extent on the premise of ensuring the performance.

## 3. Our Proposed DSP Scheme

### 3.1 Reusing of state value – Dueling Deep Q-Learning Network (RSV-DuDQN)

When Dueling DQN is used to learn hyperparameters in IDW algorithm, the convergence speed and stability after convergence of Dueling DQN are somewhat lower than other classical deep reinforcement learning algorithms, and the performance of the algorithm needs to be improved. To solve this problem, an improved DuDQN model is proposed. Reusing of State Value - Dueling Deep Q-Learning Network (RSV-DuDQN) enhances the internal relationship between state and action by combining the state value of the value function part in Q network with the reward value of the action performed in the current state, and enhances the reward signal of each state-action pair, which makes the agent more sensitive to environmental rewards when in a good state and less sensitive to rewards when in a bad state. This makes the algorithm converge faster, and after convergence, the fluctuation amplitude is greatly reduced, which improves the stability of the algorithm.

In the training of the DuDQN, the reward signal value is r, which represents the reward of the environment for the behavior after performing an action a in state s. In RSV-DuDQN, the reward value formula is shown in the following formula (9).

$$R(s,a,p) = r(s,a,p) + \lambda V(s;\theta,\beta) \quad (9)$$

Where P represents the probability that the environment will be transferred to the next state after the execution of action A under the current state S. In the hyperparameter learning of IDW algorithm, P is determined. V (s; w, θ) is the output of the value function part of the Q network. Λ is the punishment factor, which ranges to (0, 1). Its role is to determine that the reward signal R of environmental feedback plays a dominant role in the whole reward value, and to prevent the loss of sensitivity to the reward signal of environmental feedback due to the excessive state value of the value function, thus making the Q network unable to converge.

The implementation details of RSV-DuDQN model is described in Fig. 2.

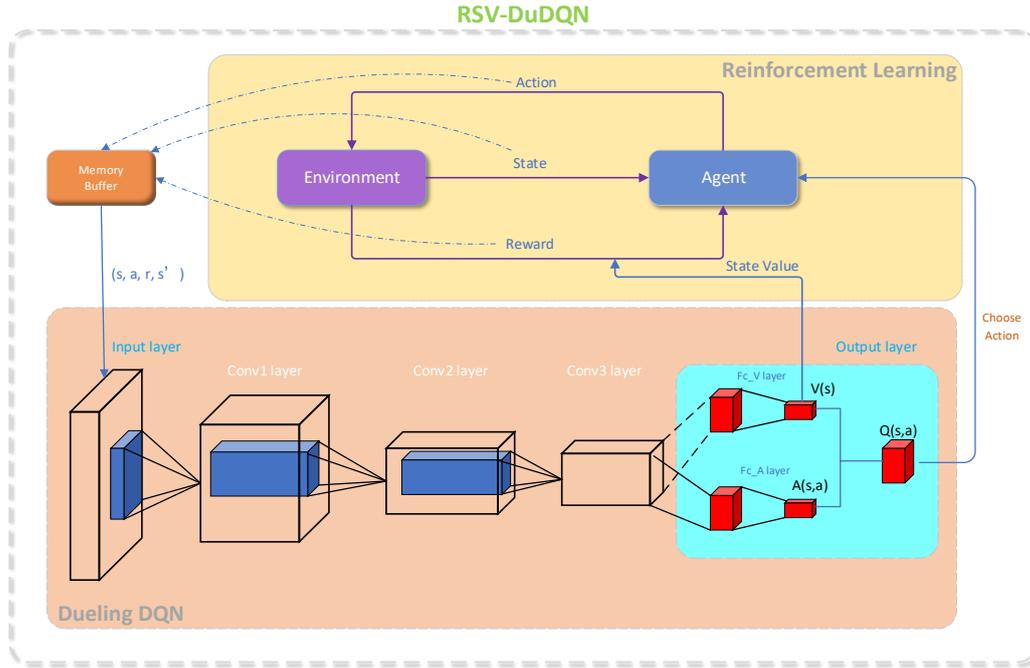

Figure 2. Schematic illustration of the RSV-DuDQN

In the actual training, the neural network will learn the direct relationship between the reward signal and the state value, and then use the method of unlimited increasing the state value of the value function to obtain a larger total environmental reward. In order to prevent this from happening, a number of training samples are extracted from the memory bank in each round of training, the state-action pairs in the samples are input and learned into the Q-network to obtain the output of the value function part, then the output is standardized, and combined with the method proposed above. The advantages are as follows:

a) By standardizing the training samples in different states, the correlation between the two adjacent states is further reduced, which is more conducive to the learning and convergence of the algorithm.
b) After standardization, the obtained status value is only related to the current status, cutting off the correlation between the total reward value and the Q network. In this way, Q network can avoid getting large total reward by directly outputting large state value.

The algorithm implementation steps are as follows:

Step 1: Initialize replay memory D to capacity N

Step 2: Initialize action- value function Q with random weights θ

Step 3: Initialize target action-value function $\hat{Q}$ with weights $\theta^- = \theta$

Step 4: For episode= 1, M do

Step 5: Initialize sequence s1={x1} and preprocessed sequenced ф1=ф（s1）

Step 6: For t = 1,T do

Step 7: With probability ε select a random action at; otherwise select at = argmaxa Q ( ф(st), a; θ )

Step 8: Execute action at in emulator and observe reward rt and image xt+1

Step 9: Set st+1 = st,at,xt+1 and preprocess ϕt+1= ϕ（st+1）

Step 10: Store transition（ϕt,at,rt,ϕt+1）in D

Step 11: Sample random minibatch of transitions（ϕj, aj, rj, ϕt+1）from D

Step 12: Standardize rj of transitions in minibatch samples to get r'j

Step 13: Input (ϕj, aj) into Q-evaluate network to get State-Value SVj

Step 14: Set Rj = r'j + λ * SVj

Step 15: If episode terminates at step j+1: yj = Rj; otherwise yj = Rj + γ maxa' $\hat{Q}$ (ϕj + 1, α'; θ⁻)

Step 16: Perform a gradient descent step on (yj − Q (ϕj; aj; θ))2 respect to network parameter θ

Step 17: Every C steps reset $\hat{Q}$ = Q

Step 18: end for

Step 19: end for

### *3.2 Proposed approach (RSV - DuDQN – IDW)*

The previous section introduces the basic theory of the IDW algorithm and the improved DuDQN model. This paper improves the accuracy of IDW algorithm by introducing RSV-DuDQN model. Because the traditional IDW algorithm does not pay attention to the structural characteristics of the interpolation space during interpolation, IDW algorithm is often unable to adapt to the complex terrain structure in the actual application process, resulting in low accuracy and large error of interpolation results. However, the parameters in the traditional IDW algorithm is related to the spatial structure characteristics, which can reflect the spatial structure characteristics of the current interpolation environment to a certain extent. In this paper, RSV-DQN model is used to train and learn the hyperparameters of each sample point in IDW interpolation, and the optimal hyperparameters corresponding to each sample are obtained. Then several spatial discrete points are obtained by combining the optimal hyperparameters with the location information of each sample point, and then the IDW algorithm is used to interpolate these spatial discrete points to get a hyperparameter distribution model. By inputting the sample points into the hyperparametric model, the hyperparameters of the IDW interpolation algorithm corresponding to the sample points can be obtained. Finally, the predicted values of the interpolation points can be obtained by IDW interpolation of the interpolation points with the obtained hyperparameters. The implementation details of RSV-DuDQN-IDW model is described in Fig. 3.

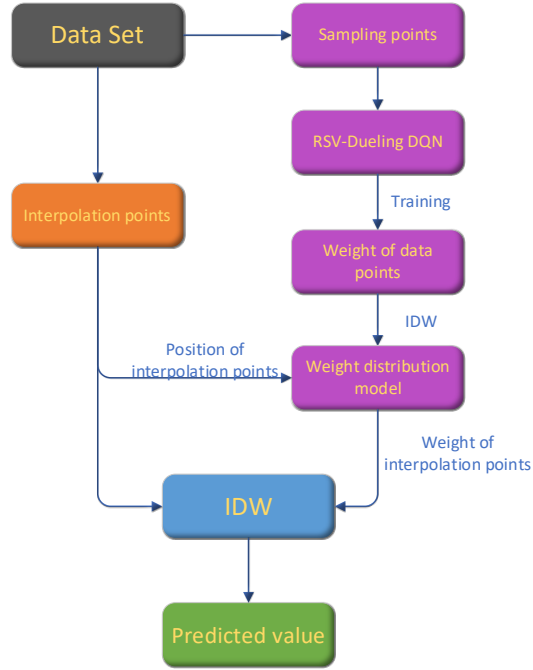

Figure 3. Schematic illustration of the RSV-DuDQN-IDW 1

## 4. Performance Analysis

In order to verify that the proposed RSV-DuDQN model has certain advantages over common deep reinforcement learning models. In Experiment 1, DQN model, DDQN model, DuDQN model and RSV-DuDQN model were used to learn the super parameters of inverse distance weighted interpolation method on common data set. In experiment 2, the hyperparameters learned from experiment 1 and the common prior parameters were used for inverse distance weighted interpolation experiment. In experiment 3, a hyperparameter distribution model is constructed by using the hyperparameters in experiment 2. According to the hyperparameter distribution model, the hyperparameters of the interpolation points can be obtained. Finally, the obtained hyperparameters are used to carry out inverse distance weighted interpolation for the interpolation points and are compared with the traditional inverse distance weighted interpolation algorithm.

The experimental environment is as follows: the processor is AMD2600, the main frequency is 3.4Ghz, and the memory is 24G. Due to the use of deep neural network in the model, matrix operation is mostly adopted, so GTX1660 graphics processor is used to carry out auxiliary acceleration operation on the model.

### *4.1 Verify the rationality of the RSV-DuDQN*

In order to verify the validity of the RSV-DuDQN model, this section uses DQN, DDQN, DuDQN and RSV-DuDQN to learn the hyperparameters of inverse distance weighted interpolation on different data sets. In all the deep reinforcement learning models, the action space of the agent is [-1, 1], and after many experiments, the action space dispersion is finally determined as [-1, -0.5, -0.1, 0, 0.1, 0.5, 1]. Through experimental certification, after the action is discretized from the continuous space to the discrete space with an accuracy of 0.1, and the influence of the hyperparameters learned by the algorithm on the entire interpolation result can be ignored.

*A. Verify the rationality of the RSV-DuDQN in Wuhan data set*

In order to verify the superiority of RSV-DuDQN model over other classical deep reinforcement learning models, this experiment uses different models to learn the hyperparameters of inverse distance interpolation algorithm on the data set of soil heavy metal content in Wuhan suburban farmland. This data set is from major technological innovation projects "Accumulation characteristics and risk assessment of heavy metals in farmland soil in Wuhan suburb " in Hubei Province. The determination method of each sample in the data set is in accordance with the requirements of the Technical Specification for Soil Environmental Monitoring (HJ/T166-2004) and Soil Environmental Quality-Risk Control standard for soil contamination of agricultural land. (GB15618-2018). The total number of sampling points is 1161, including eight common soil heavy metals, they are As, Cd, Cr, Cu, Hg, Ni, Pb and Zn. In order to conduct the experiment more rationally, four data sets with different data ranges are selected as the experimental data for this time, namely the Cd, Cr, Ni and Pb data sets. At the beginning of the experiment, ArcGIS+ software was used to convert the graticule coordinates in the original data into commonly used plane rectangular coordinates, and then standardized and input them into the Q network together with the initialized hyperparameters. The model training of the four metals is shown in Fig. 4-Fig. 7. The abscissa is the number of training steps, and the unit is (Times). The ordinate represents the error between the actual value and the predicted value obtained after interpolating the data with the currently learned hyperparameters., with the unit of (mg / kg). In order to better present the state of each algorithm during the training, all the graphs are smoothed. Table 1 shows the training of the four deep reinforcement learning models in the hyperparameter learning of data sets with different heavy metal content. The table shows the convergence time of various models in the training of data sets with different heavy metal content, the unit is (seconds), and the accuracy is 0.01. In order to show the algorithm training more intuitively, when the number of training rounds reaches 5000, the training is stopped. In this case, if the algorithm has not converged, then the convergence time in the table is uniformly expressed in the form of ">>".

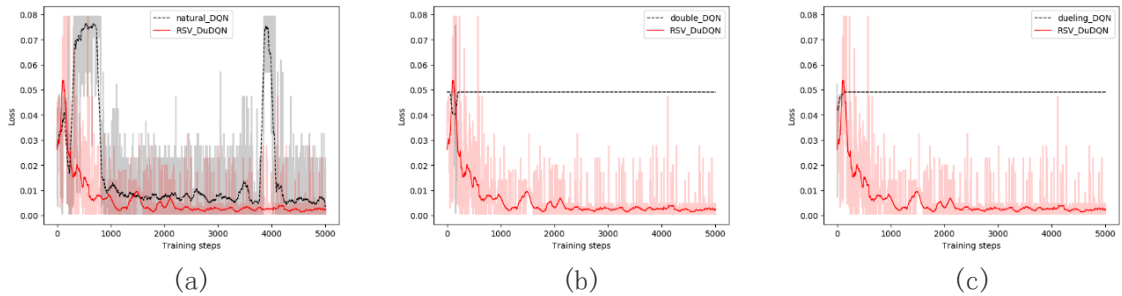

(a)  (b)  (c)

Figure 4. Training of different models on dataset Cd

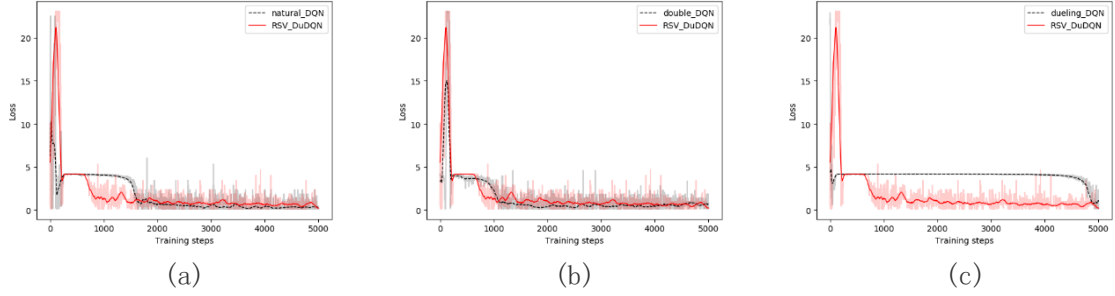

(a) (b) (c)

Figure 5. Training of different models on dataset Cr

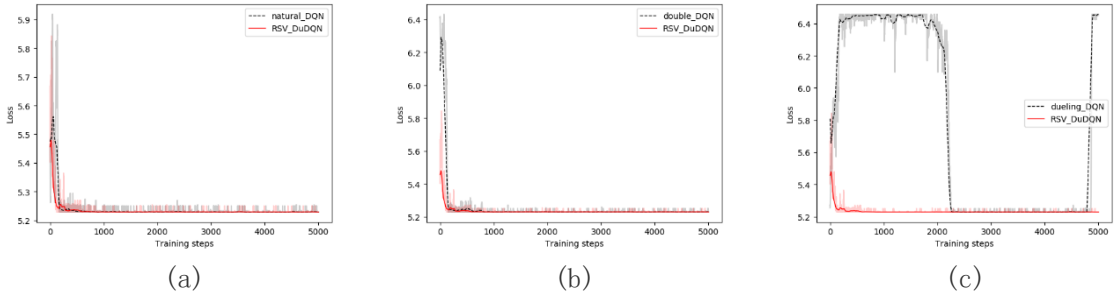

(a) (b) (c)

Figure 6. Training of different models on dataset Ni

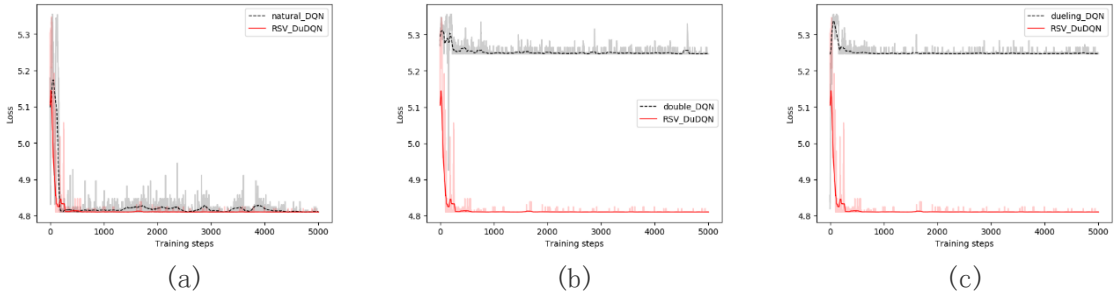

(a) (b) (c)

Figure 7. Training of different models on dataset Pb

Table 1. Convergence time of different models on Wuhan dataset

| Model | Cd | Cr | Ni | Pb |
| --- | --- | --- | --- | --- |
| DQN | 9.51 | 12.79 | 2.97 | 2.85 |
| DDQN | >> | 8.44 | 3.1 | >> |
| DuDQN | >> | 29.98 | 14.89 | >> |
| RSV-DuDQN | 8.55 | 8.09 | 2.88 | 2.54 |

As can be seen from Table 1, for different data sets, DQN, DDQN, DuDQN and RSV-DuDQN converge at different speeds. The DDQN model converges rapidly on data set Cr and Ni, while it does not converge within 5000 rounds on data set Cd and Pb, showing a poor performance. The DQN model converges on all test data sets, and it converges quickly on data set Cd, data set Ni and data set Pb. Its convergence time is second only to that of the RSV-DuDQN model. The DuDQN model performs poorly in all test data sets and even does not converge within the specified time on datasets Cd and Pb. However, the RSV-DuDQN model converges on all test data sets, and its convergence rate is higher than that of other classical deep reinforcement learning models to various degrees. It is well known that the size of interpolation data is also one of the factors affecting the convergence rate of the model. Since interpolation data are not standardized in the experiment, the four data sets represent four data ranges of different sizes. However, the RSV-DuDQN model performs better than other deep reinforcement learning models on these four data sets, indicating that the robustness of the RSV-DuDQN model is higher than that of other deep reinforcement models to some extent. In Fig. 4- Fig. 7, The initial value of each model is not consistent, because the agent will take random actions and store the information in the memory before the agent starts training. The model is trained only after enough training samples are saved in the memory bank. The initial state of each model is consistent, and the state of agent at the beginning of training is determined by the random action taken before. Therefore, it is normal that the initial value of each model is different, which does not affect the fairness of the results. In order to make the model learn the optimal hyperparameter as much as possible, the model uses greedy strategy to select actions. Therefore, after the convergence of each model, the agent does not always stay at the optimal solution, but fluctuates in a small range near the optimal solution. As can be seen from Fig. 4 and Fig. 6, DQN, DDQN and DuDQN still fluctuate greatly after the agent finds the optimal solution. They do not always stay in the optimal solution state, and sometimes even return to the poor initial state. This phenomenon is particularly evident in the performance of the DQN model in Figure 4(a). For Fig. 4- Fig. 7, the RSV-DuDQN model can always find the optimal solution quickly and stabilize around the optimal solution state. This shows that the RSV-DuDQN model has some advantages in stability compared with other models.

*B. Verify the rationality of the RSV-DuDQN in Yinchuan data set*

To reduce experimental contingency and verify that RSV-DuDQN model is more robust than DQN, DDQN and DuDQN, this experiment uses and experimented with a different public dataset, the surface soil heavy metal content dataset of Yinchuan city, Ningxia [25]. The dataset is from the Journal of Global Change Data, with a total of 96 sampling sites, including six common soil heavy metals, Co, Cr, Cs, Mg, Pb, and Ti. In order to carry out the experiment more rationally, three datasets with different data ranges are selected as the experimental data, which are Co, Mg and Pb datasets. The experimental method is the same as experiment 1. The original data are preprocessed and inputted into the deep reinforcement model for training. The training situation of all models is shown in Fig. 8- Fig. 10. The abscissa is the number of training, and the unit is (times). The ordinate represents the error between the predicted value and the true value, in units of (mg/kg), by interpolating the data using the currently learned hyperparameters. The convergence time of various models in the training of data sets with different heavy metal content is shown in Table 2, the unit is (seconds), and the accuracy is 0.01.

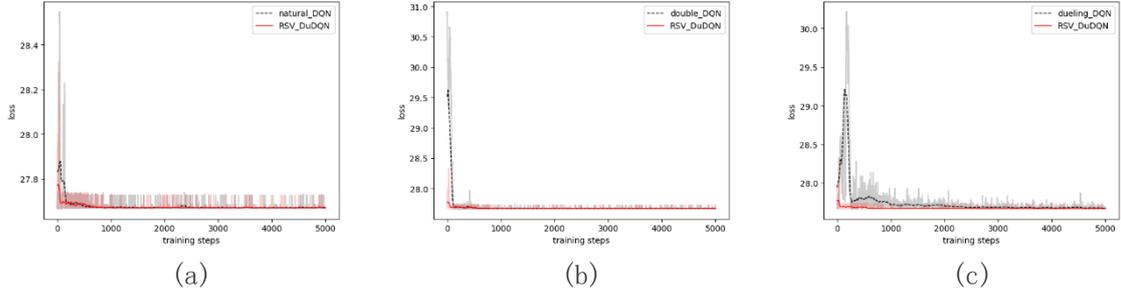

Figure 8. Training of different models on dataset Co

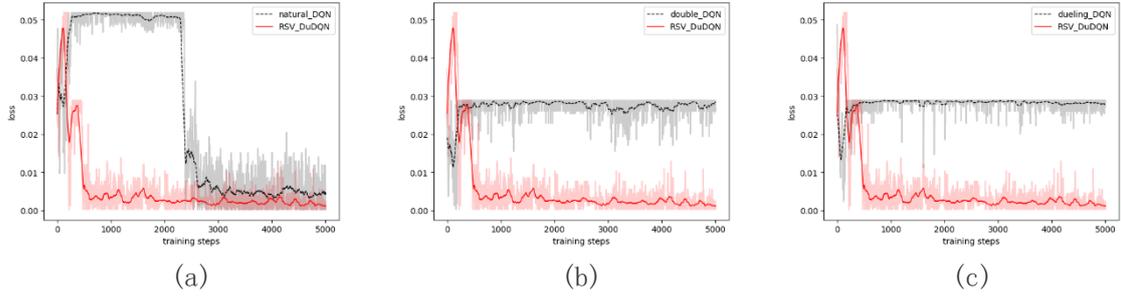

Figure 9. Training of different models on dataset Mg

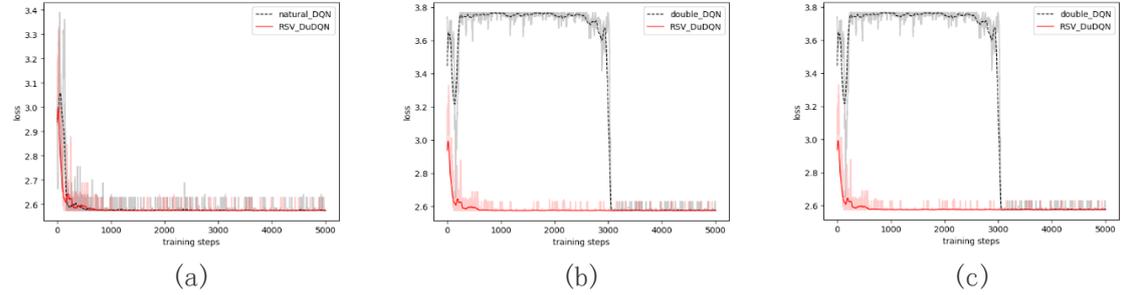

Figure 10. Training of different models on dataset Pb

Table 2. Convergence time of different models on Yinchuan dataset

| Model | Co | Mg | Pb |
| --- | --- | --- | --- |
| DQN | 1.37 | 8.59 | 1.84 |
| DDQN | 1.41 | >> | 14.66 |
| DuDQN | 4.11 | >> | 11.6 |
| RSV-DuDQN | 1.22 | 2.76 | 1.82 |

As can be seen from Table 2, the convergence time of DQN model on dataset Mg and Pb is far less than that of DDQN model and DuDQN model, while the convergence time on dataset Co is slightly less than that of DDQN model and DuDQN model. This indicates that DQN performs better than DDQN and DuDQN models in this experimental data. The convergence time of the DDQN model on the dataset Co is only slightly less than that of the DuDQN model, while the DuDQN model on the dataset Pb is slightly less than that of the DDQN model. In addition, neither the DDQN model nor the DuDQN model converges within the specified time on the dataset Mg. This indicates that the performance of DDQN model and DuDQN model is approximately the same on this dataset, and the performance gap is not obvious. The convergence time of the RSV- DuDQN model on these three data sets is less than that of other deep reinforcement learning models to varying degrees, which indicates that the RSV- DuDQN model has a certain advantage over other deep reinforcement learning models in terms of convergence rate. According to the training curve of the DQN model without smoothing in Fig. 9(a), the DQN model will get better parameters after 2500 steps of training, but after 2500 steps, the agent state of the DQN model fluctuates greatly, showing instability. It can be seen from the Fig. 9(a) and Fig. 9 (b) that the DDQN model and DuDQN model cannot learn the optimal parameter on the dataset Mg. Based on Table 2 and Fig. 8-10, it can be seen that the RSV-DuDQN model has a faster convergence rate and higher stability compared with other depth strengthening models.

**4.2 *Hyperparameter estimation based on RSV-DuDQN***

In order to verify that the hyperparameters learned by the RSV-DuDQN model are more advantageous than the commonly used prior hyperparameters, this part uses the hyperparameters learned by the RSV-DuDQN model to carry out inverse distance interpolation experiment on the data set, and at the same time uses the common prior hyperparameters to carry out comparison experiment.

*A. Hyperparameter estimation based on Wuhan data set*

In this experiment, the hyperparameters learned from the RSV-DuDQN model in 4.1(A) experiment are used for interpolation experiment. After the interpolation, the predicted value is compared with the actual value. Mean square error (MSE), mean absolute error (MAE), root mean square error (RMSE) and mean absolute percentage error (MAPE) are calculated. The common hyperparameters are treated by the same method, and the corresponding MSE, MAE, RMSE and MAPE are obtained. The error graph of the predicted value and the actual value of each sampling point is shown in Fig. 11. In the figure. The abscissa represents the sampling points in the data set, while the ordinate represents the difference between the predicted value and the actual value, with the unit of (mg/kg) and the accuracy of 0.01. The error table of the predicted value is shown in Table 3, with an accuracy of 0.01. The experiment on the Cd dataset is marked as "id =1" in the table. Experiments are carried out on Cr, Ni and Pb data in the same way, marked as "id =2", "id =3" and "id =4" respectively.

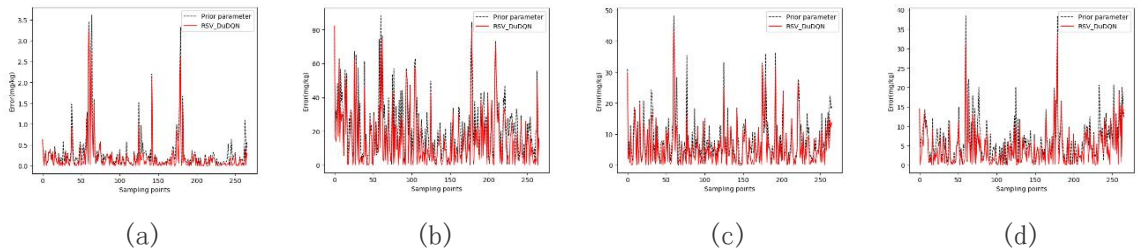

(a)　　　　　　　　(b)　　　　　　　　(c)　　　　　　　　(d)

Figure 11. Interpolation error graphs of different parameters on Wuhan dataset

Table 3. MSE, MAE, RMAE, MAPE of different parameters on Wuhan dataset

|      | MODEL           | MSE    | MAE   | RMSE  | MAPE (%) |
|------|-----------------|--------|-------|-------|----------|
| id=1 | Prior parameter | 0.34   | 0.3   | 0.58  | 461.15   |
| id=1 | RSV-DuDQN       | 0.21   | 0.19  | 0.46  | 266.02   |
| id=2 | Prior parameter | 810.56 | 22    | 28.47 | 41.94    |
| id=2 | RSV-DuDQN       | 500.54 | 14.96 | 22.37 | 29.32    |
| id=3 | Prior parameter | 123.99 | 8.09  | 11.14 | 31.67    |
| id=3 | RSV-DuDQN       | 76.25  | 5.71  | 8.73  | 22.44    |
| id=4 | Prior parameter | 69.88  | 6.21  | 8.36  | 31.78    |
| id=4 | RSV-DuDQN       | 42.34  | 4.3   | 6.51  | 22.17    |

Since the ordinate of Fig. 11 represents the difference between the actual value and the predicted value, the closer the error curve of the model in the figure is to the X axis, the more accurate the interpolation result of the model is and the better the interpolation effect is. It can be seen from Fig. 11 that the curve representing the RSV-DuDQN model is below the curve representing the prior hyperparameters at almost all sampling points. This phenomenon indicates that the hyperparameters learned from the RSV-DuDQN model are indeed superior to the common prior hyperparameters to some extent. The four different heavy metal data sets represent different data ranges. The range of data set Cd is [0.01, 4.94], and the data set is about 0.21mg/kg, which is much smaller than other data sets. Therefore, in Table 3, the mean square error and root mean square error of experiment "id =1" are much smaller than those of other experiments and the average absolute percentage error is much larger than that of other experiments. Similarly, since the data range of data set Cr is [11.13, 171.21], the data set is around 57.49mg/kg, far higher than other data sets. Therefore, in Table 3, the mean square error and root mean square error of experiment "ID =2" are much higher than those of other experiments. According to the four experiments in Table 3, the mean square error, mean absolute error, root mean square error and mean absolute percentage error based on the RSV-DuDQN model are all lower than the mean square error, mean absolute error, and mean absolute percentage error based on common hyperparameters to varying degrees. In the four experiments, the mean square error based on RSV-DuDQN model is 38.24%, 38.27%, 37.70% and 38.15% lower than that based on common super parameters respectively, indicating that the hyperparameters learned by RSV-DuDQN model are indeed effective and feasible.

B. *Hyperparameter estimation based on Yinchuan data set*

In order to reduce the experiment contingency and further verify the effectiveness of the hyperparameters learned by the RSV-DuDQN model, this experiment uses the hyperparameters learned by the RSV-DuDQN model in 4.1(B) experiment to conduct interpolation experiment. The data are processed in the same way as in the 4.2(A) experiment, and the errors are calculated. The error of the predicted value and the actual value of each sampling point is shown in Fig. 12. In the figure, the abscissa represents the sampling points in the data set, while the ordinate represents the difference between the predicted value and the actual value, with the unit of (mg/kg) and the accuracy of 0.01. The error table of the predicted value is shown in Table 4, with an accuracy of 0.01. The experiment on the Co data set is marked as "id =1" in the table. Experiments are carried out on Mg and Pb data in the same way, marked as "id =2" and "id =3" respectively.

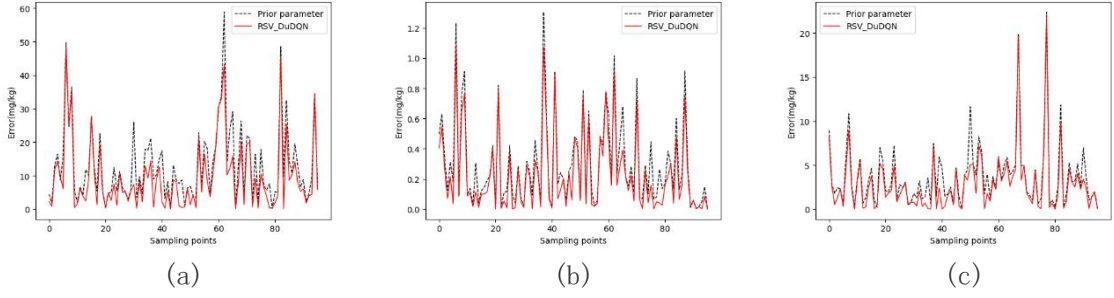

Figure 12. Interpolation error graphs of different parameters on Yinchuan dataset

Table 4. MSE, MAE, RMAE, MAPE of different parameters on Yinchuan dataset

|      | MEDEL           | MSE    | MAE   | RMSE  | MAPE (%) |
| ---- | --------------- | ------ | ----- | ----- | -------- |
| id=1 | Prior parameter | 294.09 | 12.96 | 17.14 | 37.35    |
| id=1 | RSV-DuDQN       | 199.39 | 9.51  | 9.51  | 26.74    |
| id=2 | Prior parameter | 0.17   | 0.29  | 0.41  | 16.71    |
| id=2 | RSV-DuDQN       | 0.12   | 0.23  | 0.35  | 13.39    |
| id=3 | Prior parameter | 26.41  | 3.67  | 5.13  | 15.09    |
| id=3 | RSV-DuDQN       | 19.94  | 2.84  | 4.46  | 11.61    |

It can be seen from Fig. 12(a) that at some sampling points, the solid red line representing the RSV-DuDQN model is far lower than the black dotted line representing the common prior hyperparameters, which indicates that the error of the hyperparameters learned by the RSV-DuDQN model is far less than that of the common prior hyperparameters at some sampling points. It can be seen from Fig. 12 that, at the most sampling points, the solid red line representing the RSV-DuDQN model is lower than the black dotted line representing the common prior hyperparameters to varying degrees. Only at very few points is the solid red line slightly higher than the dotted black line. This is because the agents of the RSV-DuDQN model adopt greedy strategy when taking actions, and use random method to take actions with low probability. Therefore, it is a normal case that the error of the hyperparameters based on RSV-DuDQN model is slightly higher than that of the common prior hyperparameters at individual points. In Table 4, since the data set Co is concentrated around 37.24kg/mg, which is much higher than other data sets, the mean square error and mean absolute error on data set Co are much larger than those of other data sets. On the contrary, since the data set Mg is concentrated around 2.13kg/mg, which is much smaller than other data sets, the mean square error and root mean square error on the data set Mg are much higher than those of other data sets. It can be seen from Table 4 that the interpolation errors of the hyperparameters based on the RSV-DuDQN model are smaller than those of the common hyperparameters in terms of the mean square error direction, the mean absolute error direction, the root mean square error direction and the mean absolute percentage error direction, with the average error decreasing by 28.70%. It shows that the hyperparameters learned by RSV-DuDQN model have some advantages over common prior hyperparameters.

*4.3 Interpolation experiment based on RSV- DuDQN-IDW algorithm*

To verify that the proposed RSV-DuDQN-IDW algorithm is superior to the classical inverse distance interpolation algorithm, this section uses the RSV-DuDQN model in the proposed RSV-DuDQN-IDW algorithm to learn hyperparameters from the experimental data, and then models the optimal hyperparameters to get a hyperparameter distribution model. The information of the interpolation points is input into the hyperparametric distribution model, and the corresponding hyperparameters of the interpolation points are calculated. These hyperparameters and the interpolation points are used for inverse distance weighted differential interpolation, and the predicted values of the interpolation points are obtained. The classical IDW algorithm is used for comparison experiment.

### A. Interpolation experiments of different algorithms on Wuhan data set

This experiment uses the results of experiment 4.1 (A) as experimental data, and divides the sample points in the dataset into training set and test set in a ratio of 4:1. To distinguish between the sample points in the training set and the test set, the points in the training set are recorded as sample points, and the points in the test set are recorded as interpolation points. The RSV-DuDQN-IDW algorithm and the classic IDW algorithm are used to experiment on the training set. The error between predicted and actual values for each sample point in the test set is shown in Fig. 13. In the figure, the abscissa represents the sampling points in the data set, while the ordinate represents the difference between the predicted value and the actual value, with the unit of (mg/kg) and the accuracy of 0.01. The comparison experiment on the Cd dataset is marked in the table as "id =1". Experiments are carried out on Cr, Ni and Pb data in the same way, marked as "id =2", "id =3" and "id =4" respectively. Finally, the predicted results of the two models are compared with the actual value, and the MSE, MAE, RMSE and MAPE were obtained. The experimental results are shown in Table 5, and all Error accuracy is set at 0.01.

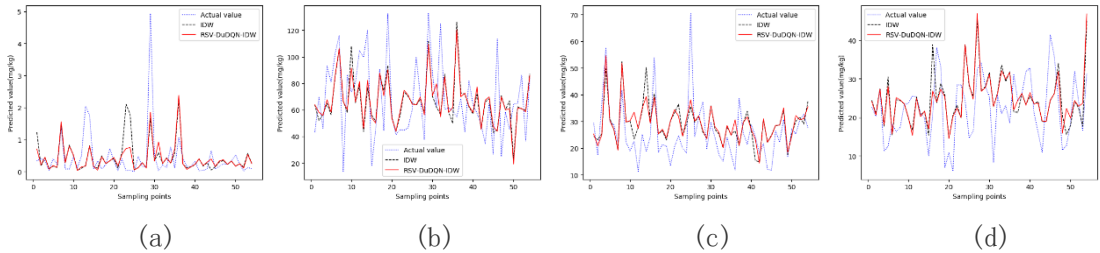

(a)　　　　　　　(b)　　　　　　　(c)　　　　　　　(d)

Figure 13. Prediction values of different algorithms on Wuhan dataset

Table 5. Interpolation errors of different algorithms on Wuhan dataset

|        | MEDEL         | MSE    | MAE   | RMSE  | MAPE (%) |
|--------|---------------|--------|-------|-------|----------|
| id=1   | IDW           | 254.84 | 12.36 | 15.96 | 41.95    |
| id=1   | RSV-DuDQN-IDW | 238.26 | 12.31 | 15.43 | 41.47    |
| id=2   | IDW           | 0.17   | 0.39  | 0.41  | 15.00    |
| id=2   | RSV-DuDQN-IDW | 0.14   | 0.28  | 0.38  | 14.38    |
| id=3   | IDW           | 19.27  | 3.49  | 4.39  | 15.31    |
| id=3   | RSV-DuDQN-IDW | 18.11  | 3.43  | 4.26  | 15.11    |

According to Fig. 13(a)(c), it can be seen that the RSV-DuDQN-IDW algorithm has a small error on dataset Cd and dataset Ni, and the predicted value is close to the actual value at most interpolation points, the predicted value and the actual value of some interpolation points are larger, but the error is still smaller than that of the classical IDW algorithm. The results show that the interpolation performance of RSV- DuDQN-IDW algorithm on Cd and Ni data sets is better than that of classical IDW algorithm. It can be seen from Fig. 13(b) (d) that the actual value of interpolation points in data sets Cr and Pb fluctuates widely and the interpolation difficulty is relatively high. Therefore, the interpolation errors of the interpolation algorithm on the dataset Cr and Pb are large. However, under such conditions, the error of RSV-DuDQN-IDW algorithm is still smaller than that of the classical IDW algorithm. It can be seen from Table 5 that the MSE of the RSV-DuDQN-IDW algorithm is smaller on dataset Cd and larger on both Cr and Pb. This is because the interpolation data range of dataset Cd is [0.01, 4.94], and that of dataset Cr and Ni are [11.13, 171.21] and [1.96, 83.30], respectively. The data value and data range of Cd are much smaller than those of Cr and Ni, which also leads to the fact that the MAPE in the interpolation experiment of dataset Cd is much larger than that of other experiments. In the interpolation experiment of data set Ni, the error value of RSV-DuDQN-IDW algorithm is only 1% smaller than that of IDW algorithm, while in the interpolation experiment of data set Cd, data set Ni and data set Pb, the MSE of RSV-DuDQN-IDW algorithm is 4.94%, 8.16% and 15.18% smaller than that of IDW algorithm respectively, indicating that the RSV-DuDQN-IDW algorithm is superior to IDW algorithm to a certain extent. In this comparison experiment, the whole interpolation process is more complex because RSV-DuDQN-IDW algorithm adds the RSV-DuDQN model. In addition to the IDW interpolation time, the RSV-DuDQN model needs extra time to learn the hyperparameters in the IDW algorithm. Obviously, the time and space complexity of the RSV-DuDQN-IDW algorithm proposed in this paper is greater than that of the traditional IDW algorithm. However, the interpolation accuracy of the RSV-DuDQN-IDW algorithm is 7.32% higher than that of the traditional IDW algorithm. Therefore, RSV-DuDQN-IDW algorithm can be applied to the fields with complex spatial structure and low requirement for real-time interpolation, but high requirements for interpolation accuracy

*B. Interpolation experiments of different algorithms on Yinchuan data set*

In order to reduce the chance of experiment and the wide applicability of RSV-DuDQN-IDW algorithm, the experimental results of experiment 4.1(B) are used as experimental data in this experiment. The experiment was performed using the same method as the 5.3 (A) experiment. The error between predicted and actual values for each sample point in the test set is shown in Fig. 14. In the figure, the abscissa represents the sampling points in the data set, while the ordinate represents the difference between the predicted value and the actual value, with the unit of (mg/kg) and the accuracy of 0.01. The comparison experiment on the Co dataset is marked in the table as "id =1". Experiments are carried out on Mg and Pb data in the same way, marked as "id =2" and "id =3" respectively. Finally, the predicted results of the two models are compared with the actual value, and the MSE, MAE, RMSE and MAPE were obtained. The experimental results are shown in Table 6, and all Error accuracy is set at 0.01.

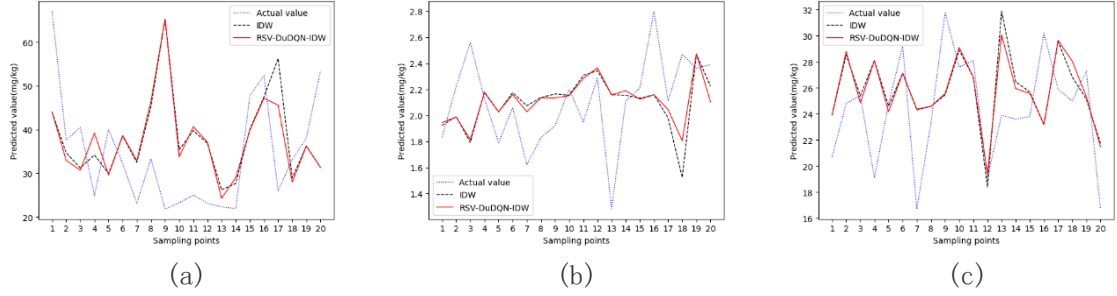

Figure 14. Prediction values of different algorithms on Yinchuan dataset

Table 6. Interpolation errors of different algorithms on Yinchuan dataset

|  | Model | MSE | MAE | RMSE | MAPE(%) |
|---|---|---|---|---|---|
| id=0 | IDW | 0.81 | 0.4 | 0.89 | 410.67 |
| id=0 | RSV-DuDQN-IDW | 0.77 | 0.38 | 0.88 | 386.59 |
| id=1 | IDW | 650.16 | 19.55 | 25.49 | 43.21 |
| id=1 | RSV-DuDQN-IDW | 597.13 | 19.51 | 24.44 | 43.39 |
| id=2 | IDW | 99.83 | 7.22 | 9.99 | 26.62 |
| id=2 | RSV-DuDQN-IDW | 98.79 | 7.24 | 9.94 | 26.48 |
| id=3 | IDW | 712.62 | 21.02 | 26.69 | 34.44 |
| id=3 | RSV-DuDQN-IDW | 604.43 | 19.81 | 24.59 | 31.79 |

As shown in Fig. 14 (a), the predictions of RSV-DuDQN-IDW algorithm and IDW algorithm deviate from the actual values of interpolation points greatly. This is due to the large range of interpolation data of Co dataset and the high fluctuation of data size at each sample point, so the interpolation on this dataset is difficult and the interpolation error is large. As shown in Fig. 14 (b) (c), the error of the RSV-DuDQN-IDW algorithm at most of the interpolation points is always less than or equal to that of the classical IDW interpolation algorithm. Table 6 shows that the RSV-DuDQN-IDW algorithm has smaller MAE and RMSE on all datasets than the IDW algorithm. Therefore, when comparing MSE of RSV-DuDQN-IDW algorithm with MSE of classic IDW, the influence of outliers can be excluded. In the interpolation experiments of dataset Co and dataset Pb, the MSE of RSV-DuDQN-IDW algorithm is 6.51% and 6.01% smaller than that of IDW algorithm, respectively, which indicates that the accuracy of RSV-DuDQN-IDW algorithm is to some extent higher than that of classical IDW algorithm. The MSE of RSV-DuDQN-IDW algorithm is 17.65% smaller than that of IDW algorithm because the size of the interpolated dataset is small and the interpolated data value of Mg is small. Combining A and B experiments, we can see that the interpolation accuracy of RSV-DuDQN-IDW algorithm is higher than that of classical IDW algorithm to some extent.

*4.4 Industrial applications*

In order to realize the scientific and sustainable development of various industries in the local industrial zone, Yinchuan municipal government sampled and analyzed the topsoil of the comprehensive industrial zone in the city. Six kinds of heavy metals which are most likely to be produced in the production process of light and textile industry, machinery, chemical industry and building materials industry were mainly detected. The six heavy metals are CS,

Mg, Pb, Co, Cr and Ti. The sampling situation is shown in Fig. 15. The purpose is to analyze the impact of various industries on the environment through the comprehensive soil pollution in various regions of the industrial zone, promote the upgrading and optimization of various industries, and realize the coordinated development of various industries.

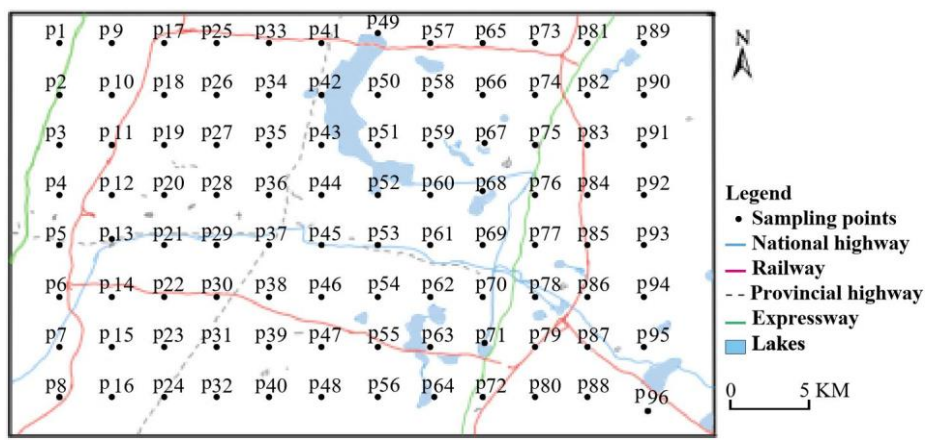

Figure 15. Sampling of the industrial zone in Yinchuan City.

## 5. Conclusion and future work

In this paper, the RSV-DuDQN model based on state value reuse is proposed and the model is fused with IDW algorithm to form the RSV-DuDQN-IDW algorithm. The RSV-DuDQN model combines the state value of the value function part in DuDQN with the reward obtained from the environmental feedback, and adds it to the reinforcement learning training in the form of total reward. The problem that DuDQN converges slowly under certain conditions and the network is still unstable after convergence is solved. In comparison with DQN, DDQN, and DuDQN, it is proved that the RSV-DuDQN model has certain advantages in convergence speed and stability. Finally, the RSV-DuDQN-IDW algorithm proposed in this paper is used to interpolate the data, and compared with the traditional IDW algorithm, the experiment proves that the RSV-DuDQN-IDW algorithm has higher accuracy and is feasible to some extent.

Although the interpolation accuracy of this algorithm is higher, the time complexity and space complexity of the algorithm are higher than the traditional IDW algorithm. For the future work, relevant methods can be used to optimize the algorithm to further reduce the time and space complexity of the algorithm while ensuring the accuracy remains unchanged.

## 6. Related work

Besides the work cited in Sections 1 and 3, there have been some other alternate mechanisms on AQM. For example, the ordinary Kriging method models and predicts the random process or random field according to the covariance function. When the Kriging method predicts the space field, it regards the space field as the generalization of the stochastic process, namely random field. In the ordinary Kriging method, the index set corresponding to random places is usually geographical coordinates, while the measurement of each point in the random field is a random variable and obeys a specific probability distribution [26]. Random fields use covariance function to describe the first law of geography, corresponding to the "kernel function" in the theory of Gaussian process regression (GPR) [27]. Using the Kriging method, the random field needs to satisfy two assumptions [28]: the first assumption is that the mathematical expectation of the random field exists and is independent of the location. The second assumption is that for any two points in the random field, the covariance function is only a function of vectors between points. However, it is not necessary to assume that the random field is an inherent stationary process when using the method proposed in this paper.

Co-Kriging is an improvement of Kriging method in dealing with multivariable problems, in which the random field to be modeled is called the main variable, and the other random fields involved in modeling are called covariates [29]. Co-Kriging can have any number of covariates, but the principal variables and covariates must have correlation and are inherently stationary random fields satisfying isotropic assumption. However, collaborative Kriging usually requires sufficient samples of covariates.

Regression Kriging is a combination of generalized linear model (GLM) and Kriging method. It is also the most common hybrid algorithm. Regression Kriging firstly uses GLM to estimate the deterministic effect in the space field, and then uses Kriging method to estimate the random field composed of regression residuals [30]. The Regression Kriging considers the trend of spatial field, so it is similar to pan Kriging, but the latter is a spatial estimation based on random field hypothesis, while regression Kriging completely separates linear trend and random process [31].

Neural network Kriging [32] is similar to regression Kriging in logic. Firstly, various artificial neural network (ANN) algorithms [33] are used to model the space field, and then Kriging method is used to estimate the residual error.

Bayes Kriging is the general name of Kriging method which uses Bayesian inference [34]. Bayes Kriging uses the prior of the hyper parameter to define the parameters (weight coefficients) of the Kriging system and estimates the posterior. Bayesian Kriging usually has normal distribution and gamma distribution [35].

Spline interpolation is a form of interpolation using a special piecewise polynomial called spline. Because spline interpolation can use low order polynomial spline to achieve small interpolation error, thus avoiding the Runge phenomenon of using higher-order polynomial [36]. This attribute enables them to meet the required smoothness constraints. Unfortunately, the same constraint often violates another ideal property: monotonicity. Monotonic cubic spline interpolation algorithm [37] simplifies and merges conditions to produce a fast method for determining monotonicity. The results are applied in the energy minimization framework to generate methods based on linear and nonlinear optimization.

The adaptive inverse distance weighted spatial interpolation algorithm can improve the performance of the interpolation algorithm by searching the " optimal" adaptive distance attenuation parameter when the spatial structure in the data cannot be effectively modeled by the typical mutation function. Regression based inverse distance weighting [38] regression inverse distance weighting algorithm improves the accuracy of the algorithm by integrating IDW and linear regression model. Pod-selective inverse distance weighting method [39] can automatically select a subset of the original set of control points, and combined with the dimension reduction technology based on the correct orthogonal decomposition of allowable displacement sets, it can reduce the calculation cost without affecting the accuracy and efficiency of interpolation.

When spatial interpolation is performed using a Multidirectional interpolation error-hiding algorithm [40], the edge classifier analyzes the value of pixels in the block around the lost block and determines which edge directions run through the lost block. Interpolate the local pixel neighborhood in the direction specified by the edge classifier. Then, use a hybrid operation to recover the missing blocks by extracting features derived from interpolation in different directions and grouping them together. When there are related pixel neighborhoods large enough, the multidirectional interpolation and image blending methods have shown very good results.

Improved RBFN (IRBF) combines a radial basis function network with an experimental semi-variogram [41]. In a radial basis network, each hidden cell has two parameters: center and width. It is generally assumed that the number of hidden cells is significantly less than the number of data points. However, for spatial interpolation methods, the number of hidden cells (neurons) is equal to the number of sample points. Radial basis networks are used to enhance the generalization ability of the model, experimental semi-variograms are used to represent spatial variability in the data, and optimal weights for the Kriging method are determined.

The Random Forest Spatial Interpolation (RFSI) model considers that nearby observations carry information about the values of a predicted location [42], and adds additional covariates to the Random Forest (RF) model, which are defined as the observations of N nearest locations and the distance from those locations to the predicted location. For each training location, n nearest locations are exported, and their observations and distances to the training locations are included as covariates along with other environmental covariates. The prediction method is the same: for each prediction location, the observations and distances from the N nearest locations are used.

When the RBF neural network [43] is used for spatial prediction, it is assumed that the sample points are independent of each other and have the same distribution characteristics. The RBF neural network method [44] has good self-learning characteristics and strong nonlinear computing ability, and overcomes the smoothing effect. Especially in the case of few sampling points, the spatial prediction results are relatively good.

**Conflict of interest statement:** All authors declare that: (a) no support, financial or otherwise, has been received from any organization that may have an interest in the submitted work; and b) there are no other relationships or activities that could appear to have influenced the submitted work.

**Author contribution statement:** J.Z designed the experiment, J.Z and C.Z performed the experiment, J.Z and N. X processed the data and wrote this paper, C.Z and N. X revised this paper.

**Acknowledgements (not compulsory):** This work was supported in part by the Hubei Provincial Major Science and Technology Special Projects under Grant 2018ABA099, in part by the Natural Science Foundation of Hubei Province under Grant 2018CFB408, in part by the National Natural Science Foundation of China under Grant 61272278, in part by the innovation and education promotion fund of science and technology development center of Ministry of education in 2019 under Grant 2018A01038,and in part by the Wuhan Polytechnic University Talent Introduction (Training) Scientific Research Project under Grant 2019RZ02.

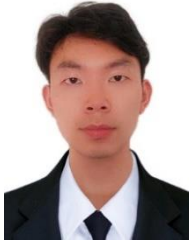
**Jun-jie Zhang** received the bachelor's degree in engineering from the Wuhan Polytechnic University, Wuhan, China, in 2018. He is currently pursuing the master's degree in mathematics and computer science with Wuhan Polytechnic University. His research interest includes Computer Vision, artificial intelligence technology and its application

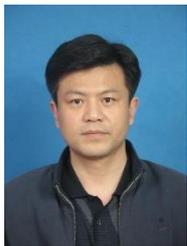
**Cong Zhang** received the bachelor's degree in automation engineering from the Huazhong University of Science and Technology, in 1993, the master's degree in computer application technology from the Wuhan University of Technology, in 1999, and the Ph.D. degree in computer application technology from Wuhan University, in 2010. He is currently a Professor with the School of Mathematics and Computer Science, Wuhan Polytechnic University. His research interests include multimedia signal processing, multimedia communication system theory and application, and pattern recognition.

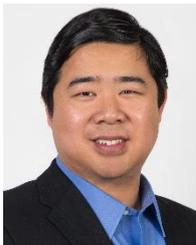
**Neal N. Xiong** is current an Associate Professor (5[rd] year) at Department of Mathematics and Computer Science, Northeastern State University, OK, USA. He received his both PhD degrees in Wuhan University (2007, about sensor system engineering), and Japan Advanced Institute of Science and Technology (2008, about dependable communication networks), respectively. Before he attended Northeastern State University, he worked in Georgia State University, Wentworth Technology Institution, and Colorado Technical University (**full professor about 5 years**) about 10 years. His research interests include Cloud Computing, Security and Dependability, Parallel and Distributed Computing, Networks, and Optimization Theory.